\PassOptionsToPackage{table}{xcolor}
\documentclass{article}
\usepackage[main,final]{neurips_2026}
\usepackage[utf8]{inputenc}
\usepackage[T1]{fontenc}
\usepackage{microtype}
\usepackage{amsmath,amssymb}
\usepackage{booktabs}
\usepackage{multirow}
\usepackage{graphicx}
\usepackage{subcaption}
\usepackage{placeins}
\usepackage{hyperref}
\hypersetup{hidelinks}
\usepackage{pgfplots}
\usepgfplotslibrary{groupplots,fillbetween}
\pgfplotsset{compat=1.18}
\usepackage{wrapfig}
\usepackage{xcolor}
\usepackage{tcolorbox}
\definecolor{promptblue}{RGB}{82,150,211}
\newtcolorbox{promptbox}[1]{
  title={#1}, colback=promptblue!8!white, colframe=promptblue,
  colbacktitle=promptblue, coltitle=white,
  fonttitle=\small\bfseries, boxrule=0.6pt, arc=2mm,
  left=3pt, right=3pt, top=3pt, bottom=3pt,
  before skip=8pt, after skip=8pt
}
\usepackage{xurl}
\usepackage{fontawesome5}
\usepackage{fancyhdr}
\graphicspath{{figures/}}
\definecolor{opdJustRL}{HTML}{0072B2}
\definecolor{opdDeepScaleR}{HTML}{D55E00}
\definecolor{opdAIME26}{HTML}{009E73}
\definecolor{opdBase}{HTML}{D55E00}
\definecolor{opdGrid}{HTML}{E3E3E3}

\pgfplotsset{
    opd axis/.style={
        axis line style={black!45},
        tick style={black!45},
        tick label style={font=\fontsize{6.5}{7.5}\selectfont},
        label style={font=\footnotesize},
        title style={font=\footnotesize\bfseries},
        legend style={font=\scriptsize, draw=none, fill=none},
        grid=major,
        grid style={draw=opdGrid, line width=0.3pt},
        axis on top,
    },
    opd initial/.style={
        dashed, mark=o, mark size=1.15pt, line width=0.7pt,
        mark options={solid, fill=white, line width=0.4pt},
    },
    opd final/.style={
        solid, mark=square, mark size=1.15pt, line width=0.7pt,
        mark options={fill=white, line width=0.4pt},
    },
    opd scatter/.style={
        only marks, mark=*, mark size=0.75pt, opacity=0.55,
    },
}

\newcommand{\westlakelogo}[1][1.10in]{\includegraphics[width=#1]{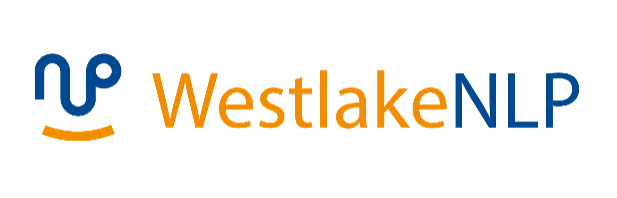}}
\AtBeginDocument{%
  \setlength{\headwidth}{\textwidth}%
  \setlength{\headheight}{39pt}%
  \addtolength{\topmargin}{-11pt}%
  \setlength{\headsep}{18pt}%
}
\fancypagestyle{westlake}{%
  \fancyhf{}
  \fancyhead[R]{\raisebox{-8pt}[\height][\depth]{\westlakelogo}}
  \fancyfoot[C]{\thepage}
  
}
\makeatletter
\renewcommand{\@notice}{}
\makeatother

\title{Gains and Collapse in On-Policy Distillation:
A Reinforcement Learning Perspective}

\author{%
  \normalfont
  \setlength{\tabcolsep}{3pt}
  \begin{tabular}{@{}ccc@{}}
    \makebox[0.31\textwidth]{\textbf{Han Cui}\thanks{Equal contribution.}} &
    \makebox[0.31\textwidth]{\textbf{Jianhao Yan}\footnotemark[1]} &
    \makebox[0.31\textwidth]{\textbf{Yun Luo}\footnotemark[1]} \\
    Zhejiang University & Westlake University & Westlake University \\
    Westlake University & & \\
    {\footnotesize\texttt{cuihan@westlake.edu.cn}} & & \\[12pt]
    \makebox[0.31\textwidth]{\textbf{Hongbo Zhang}} &
    \makebox[0.31\textwidth]{\textbf{Zhizhang Fu}} &
    \makebox[0.31\textwidth]{\textbf{Yue Zhang}\thanks{Corresponding author.}} \\
    Zhejiang University & Westlake University & Westlake University \\
    Westlake University & & {\footnotesize\texttt{zhangyue@westlake.edu.cn}}
  \end{tabular}
}

\newcommand{\passk}[1]{\textsc{pass}@#1}

\begin{document}

\maketitle
\vspace{-12pt}
{\centering\small
  \href{https://github.com/HancCui/opd_hacking}{\textcolor[rgb]{0.19,0.36,0.47}{\faGithub\enspace:\enspace\nolinkurl{https://github.com/HancCui/opd_hacking}}}\\[3pt]
  \href{https://hanccui.github.io/opd_hacking}{\textcolor[rgb]{0.19,0.36,0.47}{\faGlobe\enspace:\enspace\nolinkurl{https://hanccui.github.io/opd_hacking}}}
  \par
}
\vspace{3pt}

\begin{abstract}
On-policy distillation (OPD) has become an important approach to language model post-training.
However, despite its performance gains, OPD can also collapse into excessively long and repetitive generation, and the mechanism underlying these divergent outcomes remains poorly understood.
We explain these outcomes through a reinforcement learning perspective: the teacher implicitly rewards student behaviors, even those it rarely exhibits itself.
From this perspective, our experiments show that OPD improves performance without expanding the student’s capabilities.
When the implicit reward model is reliable, OPD makes correct responses easier to sample.
In contrast, when the preference misaligns with quality, reward hacking happens:
the implicit reward model amplifies overlong, repetitive student rollouts, even though it rarely generates such text itself.
Guided by this diagnosis, we find that masking unhealthy responses during training and using SFT initialization can each effectively mitigate the collapse.
Together, these findings show that OPD amplifies student behaviors favored by the teacher's implicit feedback, shifting the focus from how well the teacher generates to how reliably it evaluates student rollouts.
Our code is available at \url{https://github.com/HancCui/opd_hacking}.

\end{abstract}

\begingroup
\setlength{\parindent}{0pt}
\noindent\includegraphics[width=\textwidth]{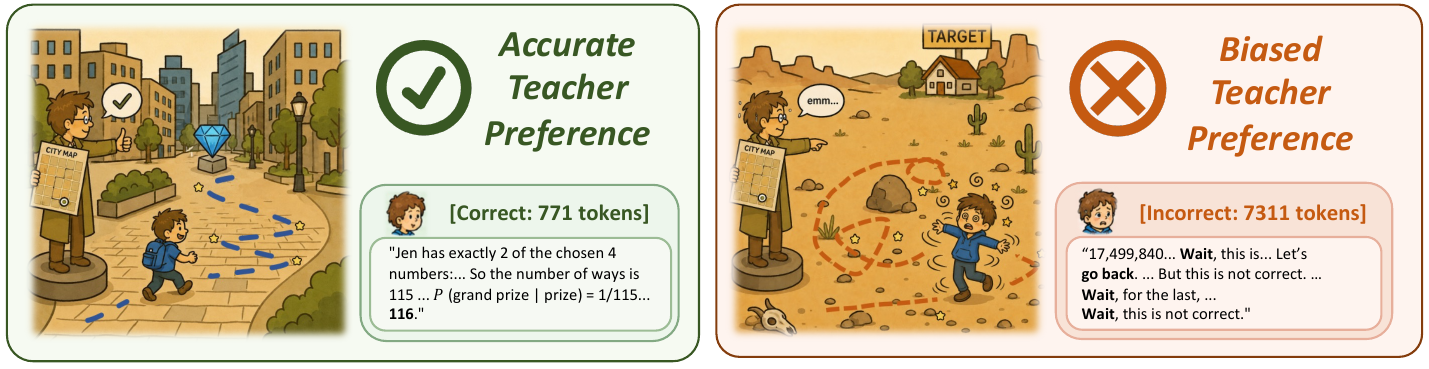}

\captionof{figure}{%
    \textbf{OPD amplifies what the student already samples.}
    The teacher scores student-generated trajectories rather than supplying
    training trajectories of its own; the examples illustrate how its preferences
    can amplify either coherent or degenerate student behavior.%
}
\label{fig:main_teaser}
\endgroup

\section{Introduction}

On-policy distillation (OPD) has become an important approach for developing frontier language models and has been adopted in large-scale post-training pipelines~\citep{deepseekai2026deepseekv4highlyefficientmilliontoken, team2026kimi, xiao2026mimo, yang2025qwen3technicalreport, zeng2026glm}.
At each iteration, the current student samples its own responses, and the teacher supervises it through token-level KL divergence at student-generated prefixes~\citep{agarwal2023gkd,gu2024minillm}.
While prior work has demonstrated the potential of OPD to improve student performance, recent studies have reported unstable training and degenerate behaviors, including excessive response length and repetition~\citep{li2026rethinkingopd,zhu2026facesonpolicydistillationpitfalls}.
It remains unclear what OPD actually learns during training, and why this learning process can sometimes lead to collapse.

In this paper, we view OPD from an RL perspective, in which the teacher implicitly rewards the student's own responses.
We investigate OPD on mathematical tasks using different types of teachers, including RL-trained reasoning models and larger general-purpose models.
Our experiments show that
\textbf{OPD does not directly extend the student's capability. Instead, it changes the relative probability of responses the student already produces according to the teacher's preferences.} 
When these preferences align with response quality, this reweighting can improve sampling efficiency; 
when they do not, it can lead to reward hacking, with the student increasingly producing responses that receive high implicit rewards despite poor quality. 
We substantiate this view through three findings.

First, we show that successful OPD reinforces correct responses the student can already generate.
We compare the initial and trained students across sampling budgets, combining pass@$k$ evaluation with additional sampling and manual audits.
Although OPD substantially improves single-sample accuracy(pass@1), the gains progressively diminish as the sampling budget increases.
At the same time, we find no expansion of the observed solvable set: problems solved after OPD can also be solved by the initial student given sufficient sampling.
Together, these results suggest that OPD makes existing capabilities easier to elicit rather than expanding the student model's problem-solving capabilities.

In contrast, we find that the observed OPD collapse reflects reward hacking: training reinforces poor-quality responses favored by the teacher.
We compare a successful OPD run with a reproduction of the collapsed setting reported by~\citet{li2026rethinkingopd}.
Despite their sharply different outcomes, both runs converge normally and concentrate on responses with low negative log-likelihood under the initial student, suggesting that collapse is not simply a failure of optimization or convergence.
In the collapsed run, the student produces increasingly overlong and repetitive responses.
At the same time, its response distribution shifts markedly toward responses preferred by the teacher over those the teacher disfavors, indicating that the student is fitting the teacher's preference signal increasingly well.
The student thus learns the teacher's preferences even as its generation quality deteriorates.

Building on this diagnosis, we mitigate OPD collapse with a standard remedy for reward hacking: limiting the reinforcement of pathological responses.
Before training, higher implicit rewards favor correct and concise responses in the successful setting, but overlong and repetitive responses in the collapsed setting, with both types of behavior later amplified during training.
Notably, the teacher itself rarely generates such pathological responses, showing that generation quality does not reliably reflect feedback quality on student rollouts.
Keeping the teacher fixed, masking rollouts that reach the generation limit mitigates degeneration and improves average evaluation accuracy by 3.08 percentage points, 
and SFT warmup provides a complementary solution by improving the quality of the student's initial rollout distribution.
Together, these results show that collapse can be mitigated by controlling which student rollouts are available for reinforcement.

Our contributions are summarized as follows:
\begingroup
\setlength{\topsep}{4pt}
\setlength{\partopsep}{0pt}
\setlength{\itemsep}{4pt}
\setlength{\parsep}{0pt}
\begin{itemize}
    \item We interpret OPD as an RL-like process with the teacher acting as an implicit reward model, linking its performance gains to improved sampling efficiency.
    \item We identify reward hacking in the observed collapse: the student fits the teacher's preferences increasingly well even as its generation quality deteriorates.
    \item We trace the collapse to a biased teacher reward signal. Limiting the reinforcement of this biased signal through masking, or reducing exposure to pathological rollouts through SFT warmup, mitigates collapse.
\end{itemize}
\endgroup

\section{On-policy distillation as implicit reward optimization}
\label{sec:opd-implicit-reward}

We formulate the sampled-token reverse-KL variant of on-policy distillation (OPD) studied in this paper as policy optimization with a teacher-derived reward.
This formulation makes explicit what the teacher scores, how those scores guide student updates, and why their reliability must be examined on student-generated trajectories.

\subsection{On-policy teacher supervision}
\label{sec:opd-supervision}

Let $q\sim\mathcal{D}$ be a training question with prompt $x_q$. We denote the fixed teacher by $T$, the initial student by $S_{\mathrm{INIT}}$, the trainable student by $S_\theta$, and the final trained student by $S_{\mathrm{OPD}}$. At each rollout iteration, the current student policy $\bar S=S_{\bar\theta}$ generates a response $y=(y_1,\ldots,y_L)$. The teacher evaluates tokens conditioned on the resulting prefixes $h_t=(x_q,y_{<t})$.

Let $d^{\bar S}$ denote the prefix distribution induced by the rollout policy. The conditional reverse-KL objective is
\begin{equation}
\mathcal{L}_T(\theta;\bar S)
=
\mathbb{E}_{h\sim d^{\bar S}}\!\left[
D_{\mathrm{KL}}\!\left(
\pi_{S_\theta}(\cdot\mid h)
\,\Vert\,
\pi_T^{(\tau_T)}(\cdot\mid h)
\right)
\right],
\label{eq:opd-conditional-kl}
\end{equation}
where $\tau_T$ is the teacher scoring temperature. We hold the rollout-prefix distribution fixed during each optimization step and refresh it through subsequent student rollouts. The student thus determines where supervision is evaluated, while the teacher supplies the target token probabilities at those states.

\subsection{The teacher as an implicit reward model}
\label{sec:opd-reward}

\paragraph{Reward interpretation.}
By expanding the KL divergence in Equation~\eqref{eq:opd-conditional-kl}, we can recast the objective into a maximum-entropy policy objective over sampled prefixes in Equation~\ref{eq:opd-reward-entropy}:
\begin{equation}
\begin{aligned}\mathcal{L}_T(\theta;\bar S)
&=- \mathbb{E}_{h\sim d^{\bar S}}\!\left[
\mathbb{E}_{a\sim\pi_{S_\theta}(\cdot\mid h)}
\bigl[r_T(h,a)\bigr]
\right.\left.
+\mathcal{H}\!\left(\pi_{S_\theta}(\cdot\mid h)\right)
\right],
\end{aligned}
\label{eq:opd-reward-entropy}
\end{equation}
where \(r_T(h,a)=\log \pi_T^{(\tau_T)}(a\mid h)\) and \(H\) denotes entropy.
Minimizing the conditional reverse KL is therefore equivalent to maximizing the expected teacher reward with entropy regularization on the sampled prefix distribution. This is the token-level reward interpretation underlying policy-gradient implementations of OPD~\citep{lu2025onpolicydistillation}.

\paragraph{Policy-gradient update.}
The reward interpretation also yields the sampled-token update. For each rollout token, define
\begin{equation}
A_t^T
=r_T(h_t,y_t)-\log\pi_{\bar S}(y_t\mid h_t).
\label{eq:opd-teacher-advantage}
\end{equation}
Writing $A^T(h,a)=r_T(h,a)-\log\pi_{\bar S}(a\mid h)$, the unclipped estimator at a fixed prefix $h$ satisfies
\begin{equation}
\begin{aligned}
&\left.
\mathbb{E}_{a\sim\pi_{\bar S}(\cdot\mid h)}\!\left[
A^T(h,a)
\nabla_\theta\log\pi_{S_\theta}(a\mid h)
\right]\right|_{\theta=\bar\theta}
\\
&\qquad=
-\left.\nabla_\theta
D_{\mathrm{KL}}\!\left(
\pi_{S_\theta}(\cdot\mid h)
\,\Vert\,
\pi_T^{(\tau_T)}(\cdot\mid h)
\right)\right|_{\theta=\bar\theta}.
\end{aligned}
\label{eq:opd-gradient-identity}
\end{equation}
In practice, we treat $A_t^T$ as a stop-gradient advantage in a PPO-style clipped surrogate and add reference-KL regularization toward $S_{\mathrm{INIT}}$. The teacher score $r_T$ and the update weight $A_t^T$ are distinct: the latter includes the student log-probability term arising from entropy regularization. These updates use immediate token-level signals on student-generated prefixes.
When most advantages are negative, the main effect is to reduce token probabilities by different amounts.
More negative advantages lead to stronger pressure to reduce the probability, while less negative advantages lead to weaker pressure.

\begin{table}[t]
\centering
\small
\caption{OPD settings used in our analysis. Each setting is named after its teacher.}
\label{tab:opd-settings}
\begin{tabular}{@{}p{0.28\linewidth}p{0.39\linewidth}p{0.25\linewidth}@{}}
\toprule
Teacher & Student initialization & Training data \\
\midrule
JustRL-1.5B & DeepSeek-R1-Distill-Qwen-1.5B & DeepMath-103K ($\geq6$) \\
DeepScaleR-1.5B-Preview & DeepSeek-R1-Distill-Qwen-1.5B & DeepMath-103K ($\geq6$) \\
Qwen3-4B & Qwen3-1.7B-Base & OpenThoughts3 \\
\bottomrule
\end{tabular}
\end{table}

\subsection{Experimental settings}
\label{sec:opd-experimental-settings}

We study three teacher--student settings, summarized in Table~\ref{tab:opd-settings}.
The JustRL and DeepScaleR settings share the same student initialization and training data.
Their teachers, JustRL-1.5B~\citep{he2025justrlscaling15bllm} and DeepScaleR-1.5B-Preview~\citep{deepscaler2025}, were obtained through RL from DeepSeek-R1-Distill-Qwen-1.5B~\citep{guo2025deepseek}, which also initializes our student. Both settings use the level $\geq6$ subset of DeepMath-103K and a 16,384-token rollout limit.
The Qwen3-4B setting follows~\citet{li2026rethinkingopd}, using Qwen3-4B to supervise Qwen3-1.7B-Base on OpenThoughts3  prompts ~\citep{guha2025openthoughtsdatarecipesreasoning} with an 8,192-token rollout limit. The teacher remains fixed in all settings.
We train for 100 rollout iterations using the Slime training framework~\citep{slime_github}, with a reference-KL coefficient of $0.001$. We evaluate mathematical reasoning on AIME24--26, reporting mean@4 for routine evaluation and using 256 responses per problem for the coverage and sampling efficiency analyses. Full training and evaluation protocols are provided in Appendix~\ref{app:training_details}.

\section{Does OPD Expand Student's Capability Boundary?}
\label{sec:opd_phenomena}

The implicit-reward perspective suggests that successful OPD can improve performance by reinforcing correct student responses.
We examine this successful case: does teacher feedback make problems already solvable by the initial student easier to solve?
We compare the initial student \(S_{\mathrm{INIT}}\) with the trained student \(S_{\mathrm{OPD}}\) using \(\mathrm{pass}@k\) across increasing sampling budgets, audited problem coverage, and per-problem success rates.

\paragraph{Gains diminish as the sampling budget grows.}
\label{sec:3_1}
\begin{figure*}[t]
    \centering
    \includegraphics[width=0.98\textwidth]{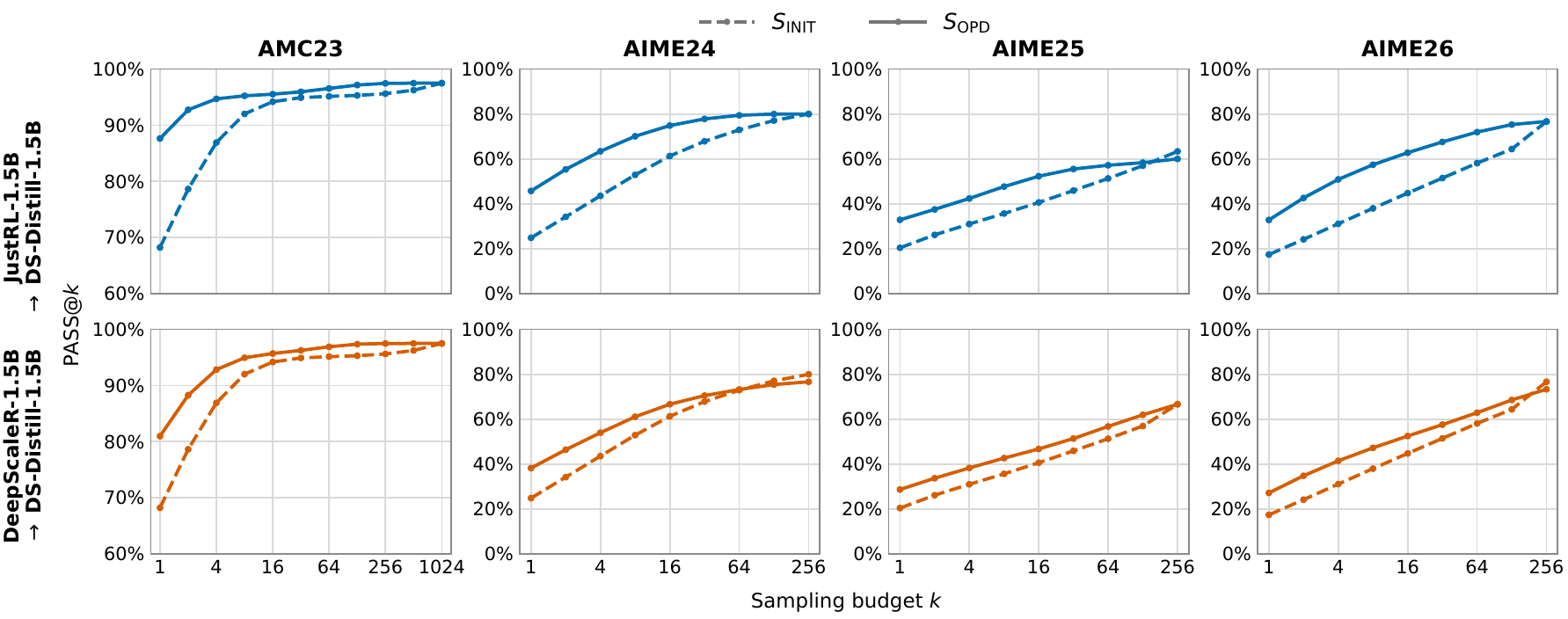}
    \caption{
        \textbf{\passk{k} curves of \(S_{\mathrm{INIT}}\) and \(S_{\mathrm{OPD}}\).}
        Across AMC23 and AIME24--26, OPD gains are largest at small \(k\) and
        shrink as the sampling budget grows. AIME curves extend to \(k=256\),
        while AMC23 extends to \(k=1024\).
    }
    \label{fig:s3_sampling_accuracy}
\end{figure*}

We compare the initial student \(S_{\mathrm{INIT}}\) with the trained student \(S_{\mathrm{OPD}}\) under both teachers, JustRL-1.5B and DeepScaleR-1.5B-Preview, on AIME24--26 and AMC23.
We compute \passk{k} from each response pool for \(k\in\{1,2,4,8,16,32,64,128,256\}\) on AIME and additionally \(k\in\{512,1024\}\) on AMC23, averaged across all test cases.

As shown in Figure~\ref{fig:s3_sampling_accuracy}, we observe a consistent trend under both teachers.
When \(k\) is small (e.g., \(k=1\), equivalent to average-case accuracy), \(S_{\mathrm{OPD}}\) clearly outperforms \(S_{\mathrm{INIT}}\).
However, as \(k\) increases, \(S_{\mathrm{INIT}}\) steadily catches up.
At the maximum available \(k\), \passk{k} equals the observed coverage: the fraction of problems solved at least once in the response pool.
This suggests that OPD improves accuracy under a small sampling budget, but does not improve the observed coverage at higher \(k\).

\paragraph{No Observed Expansion in Problem Coverage After OPD.}
To distinguish new coverage from finite-sample misses, we first identify the problems solved by $S_{\mathrm{OPD}}$ but not by $S_{\mathrm{INIT}}$ within the original 256-response pools of AIME24-26.
For each of these problems, we draw 768 additional responses from \(S_{\mathrm{INIT}}\), bringing its pool to 1024 responses.
Manual review checks derivation validity and corrects verifier misses in the existing outputs.

\begin{wrapfigure}{r}{0.5\textwidth}
    \centering
    \vspace{-5mm}
    \includegraphics[width=0.85\linewidth]{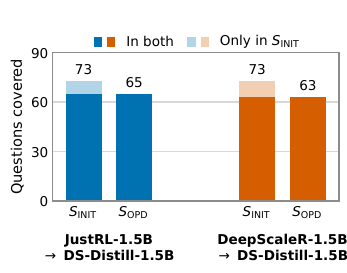}
\captionsetup{font=footnotesize}
    \caption{
        \textbf{Change in observed coverage from \(S_{\mathrm{INIT}}\) to \(S_{\mathrm{OPD}}\).}
        No problem is solvable by \(S_{\mathrm{OPD}}\) but not by \(S_{\mathrm{INIT}}\) in our observation.
        Counts come from the audited pools with conditional extension to 1,024
        responses.
    }
    \label{fig:s3_coverage_partitions}
    \vspace{-7mm}
\end{wrapfigure}

After additional sampling and manual review, no problem remains solved only by
$S_{\mathrm{OPD}}$: candidate cases are resolved either by finding a valid
$S_{\mathrm{INIT}}$ solution or by excluding an invalid $S_{\mathrm{OPD}}$
derivation. The audited $S_{\mathrm{INIT}}$ set contains 73 solvable problems,
against which $S_{\mathrm{OPD}}$ covers 65 under JustRL-1.5B and 63 under
DeepScaleR-1.5B-Preview. These counts are larger than the $k=256$ endpoints of
Figure~\ref{fig:s3_sampling_accuracy}, which are computed before the audit.
Detailed coverage analyses are provided in Appendix~\ref{app:coverage_records}.
These indicate that OPD does not induce an observed expansion of the solvable set.

\paragraph{Already-solvable problems become easier to solve.}
\label{sec:3_3}

\begin{wrapfigure}{r}{0.5\textwidth}
    \centering
    \vspace{-5mm}
    \includegraphics[width=\linewidth]{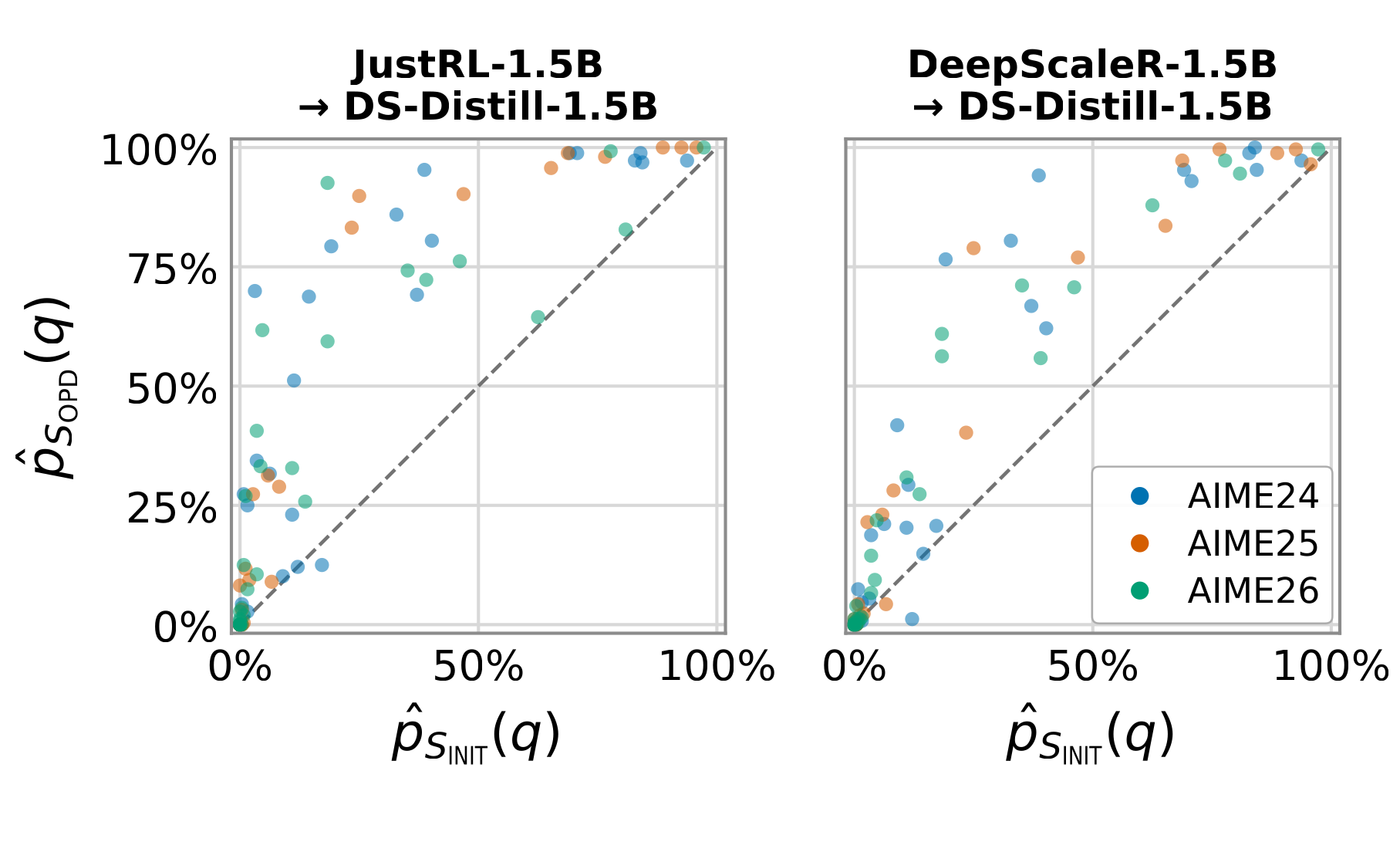}
\captionsetup{font=footnotesize}
    \caption{
        \textbf{Per-problem mean@256 of \(S_{\mathrm{INIT}}\) ($x$-axis) and \(S_{\mathrm{OPD}}\) ($y$-axis).}
        OPD raises success rates on problems already solvable by \(S_{\mathrm{INIT}}\).
        The dashed line marks equal score.
    }
    \label{fig:s3_sampling_efficiency}
    \vspace{-5mm}
\end{wrapfigure}

To identify the source of OPD's small-\(k\) gains, we compare per-problem success rates between \(S_{\mathrm{INIT}}\) and \(S_{\mathrm{OPD}}\).
For each problem \(q\), let \(\hat{p}_m(q)\) denote the fraction of correct responses among the 256 samples from model \(m\).

As shown in Figure~\ref{fig:s3_sampling_efficiency}, the problem-level distribution is strongly shifted above the diagonal under both teachers.
OPD increases the per-problem success rate for 90.6\% of the problems already solvable by $S_{\mathrm{INIT}}$ with JustRL-1.5B as the teacher, and for 84.4\% with DeepScaleR-1.5B-Preview.
Averaged over these problems, the increase in mean@256 is 22.6\% and 14.7\%, respectively.

Problems unimproved by OPD are those that \(S_{\mathrm{INIT}}\) could hardly solve, and no high-performing problem drops to near-zero accuracy (lower-right corner).
Some even jump from low to high success rates (upper-left corner), showing that OPD can make problems the initial student rarely solves much easier to solve.
Thus, OPD primarily eases sampling of correct answers on already-solvable problems, which accounts for the small-k performance gain.

\paragraph{Summary.}
In the two successful settings studied here, OPD improves sampling efficiency on problems already solvable by the initial student.
Its positive accuracy gains disappear at the largest sampling budgets tested, and no validated OPD-only problem remains after the targeted AIME coverage audit.
These findings are consistent with the teacher acting as an implicit reward model whose feedback reinforces correct student behavior, making successful responses easier to sample.

\FloatBarrier
\section{Reward Hacking in OPD: Convergence to a Biased Reward Signal}

\begin{figure}[t]
    \centering
    \includegraphics[width=0.95\textwidth]{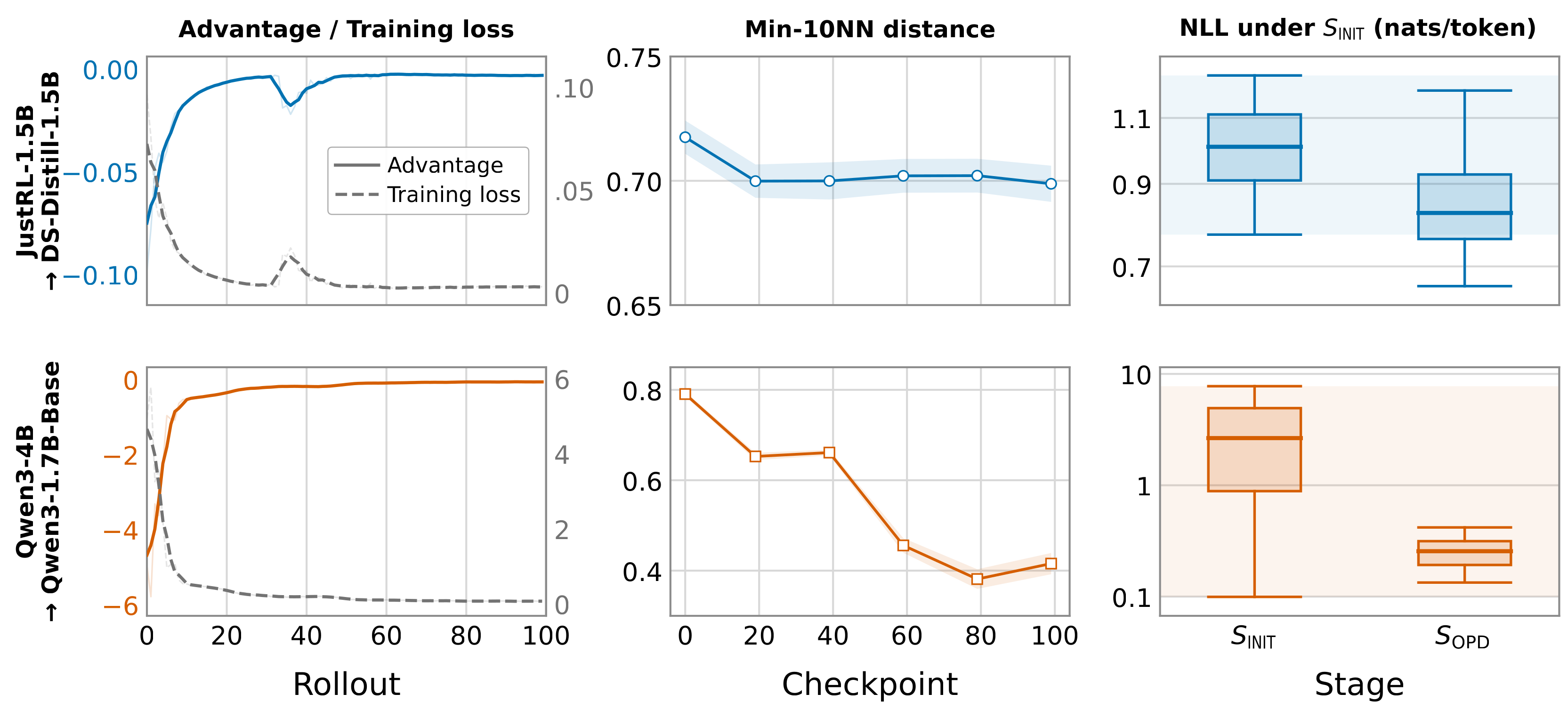}
    \caption{
        \textbf{Training process of the JustRL (top) and Qwen3-4B (bottom) settings.}
        Optimization converges normally in both settings, and the trained student's responses remain highly likely under $S_{\mathrm{INIT}}$.
        Columns: the advantage and training loss; the Min-10NN distance over six checkpoints; the NLL under $S_{\mathrm{INIT}}$ of responses from $S_{\mathrm{INIT}}$ and $S_{\mathrm{OPD}}$.
    }
    \label{figure_4_1}
\end{figure}

The preceding analysis shows that OPD improves performance without expanding the set of problems the student can solve.
Yet OPD training does not always succeed: some teacher-student pairs produce collapsed behaviors instead.
From a diagnostic perspective, we then examine whether the collapse constitutes reward hacking: optimization succeeding at the wrong target.
We first verify that the optimization converges normally, then characterize the behavioral shifts that emerge during training.

\subsection{Collapsed OPD Training Still Converges Normally}
\label{sec:4_1}

Our analysis starts by comparing a reported failure case of OPD with a successful one.
Prior work~\citep{li2026rethinkingopd} reports that OPD can fail when the teacher and student distributions are substantially mismatched, attributing the resulting degradation to this large distribution gap.
We therefore compare the full training process of the two settings:
the \textbf{JustRL setting} (JustRL-1.5B $\rightarrow$ DS-Distill-1.5B) where training improves the student,
and the \textbf{Qwen3-4B setting} (Qwen3-4B $\rightarrow$ Qwen3-1.7B-Base, without warm-up) where training collapses.
The loss and advantage in the Qwen3-4B setting converge as smoothly as those in the JustRL setting, with no qualitative difference in the shape of the curves.

We measure response concentration using the Min-10NN distance~\citep{zhang2026detectingrlvrtrainingdata}, which averages the ten smallest nearest-neighbor distances among responses to the same problem; lower values indicate greater similarity (described in detail in Appendix~\ref{app:response_geometry}).
The distance keeps decreasing during training in both settings: mildly in the JustRL setting, and much more sharply in the Qwen3-4B one.
Both students shrink toward a narrower space of behaviors, a phenomenon also reported in other work on on-policy RLVR post-training~\citep{yue2025does, zhao2025echochamberrlposttraining, dang2025assessing}.

If training exploits the reward rather than learning from it, the selected behaviors should already be accessible to $S_{\mathrm{INIT}}$ rather than genuinely novel.
To examine this, we compute the NLL under $S_{\mathrm{INIT}}$ for responses generated by the trained student and compare it with the NLL of $S_{\mathrm{INIT}}$'s own responses to the same problems.
Figure~\ref{figure_4_1} summarizes these diagnostics; the right panel reports this NLL comparison.
In both settings, the trained student's responses lie in the low NLL region of $S_{\mathrm{INIT}}$, which indicates that they are already familiar to $S_{\mathrm{INIT}}$.

\subsection{The Student Faithfully Fits the Reward Signal, Even When the Teacher Is Hacked}
\label{sec:4_2}

If the collapse constitutes reward hacking, the trained student should have successfully internalized the teacher's preferences, but those preferences should be misaligned with solution quality.

\begin{wraptable}{r}{0.48\textwidth}
    \centering
    \captionsetup{font=small}
    \begingroup
    \centering
    \scriptsize
    \setlength{\tabcolsep}{1.5pt}
    \vspace{-3mm}
    \renewcommand{\arraystretch}{1.08}

    \caption{
        \textbf{Endpoint generation characteristics on AIME24--26.}
        Model pairs denote teacher $\rightarrow$ student.
        Bold values mark length-limit saturation and severe repetition.
    }
    \label{tab:endpoint_generation_characteristics}

    \begin{tabular*}{\linewidth}{@{\extracolsep{\fill}}lrrrr@{}}
        \toprule
        Model & Acc. (\%) & Len. (k) & Trunc. (\%) & Rep. (\%) \\
        \midrule

        \rowcolor{black!6}
        \multicolumn{5}{@{}l}{\rule{0pt}{2.2ex}\textbf{JustRL-1.5B $\rightarrow$ DS-Distill-1.5B}} \\[1pt]
        Teacher & 39.3 & 9.30 & 13.8 & 0.0 \\

        $S_{\mathrm{INIT}}$ & 20.9 & 11.19 & 12.5 & 0.0 \\

        $S_{\mathrm{OPD}}$ & 37.1\,{\tiny(+16.2)} & 9.72 & 16.5 & 0.0 \\

        \midrule

        \rowcolor{black!6}
        \multicolumn{5}{@{}l}{\rule{0pt}{2.2ex}\textbf{Qwen3-4B $\rightarrow$ Qwen3-1.7B-Base}} \\[1pt]
        Teacher & 19.6 & 3.09 & 2.1 & 0.0 \\

        $S_{\mathrm{INIT}}$ & 0.2 & 1.18 & 4.1 & 4.2 \\

        $S_{\mathrm{OPD}}$ & 5.4\,{\tiny(+5.2)} & \textbf{8.16} & \textbf{99.4} & \textbf{38.0} \\

        \bottomrule
    \end{tabular*}
\endgroup

\end{wraptable}

\textbf{Endpoint behaviors are pathological.}
Table~\ref{tab:endpoint_generation_characteristics} compares the free generations of the teacher, $S_{\mathrm{INIT}}$, and $S_{\mathrm{OPD}}$ across both settings.
In the JustRL setting, the trained student's behaviors closely match the teacher's: accuracy improves by 16.2 points, and no severe repetition is detected.
In contrast, the Qwen3-4B setting exhibits a dramatically different outcome.
Although accuracy improves modestly (+5.2 points), nearly all responses hit the 8k generation limit (99.4\%), and 38\% exhibit severe repetition---far worse than either the teacher (2.1\% truncation, 0\% repetition) or $S_{\mathrm{INIT}}$ itself.
The trained student has not learned to reason like the teacher; instead, it has concentrated probability mass on a narrow subset of pathological behaviors.

\clearpage

\begin{wrapfigure}{r}{0.48\textwidth}
    \centering
    \includegraphics[width=0.9\linewidth]{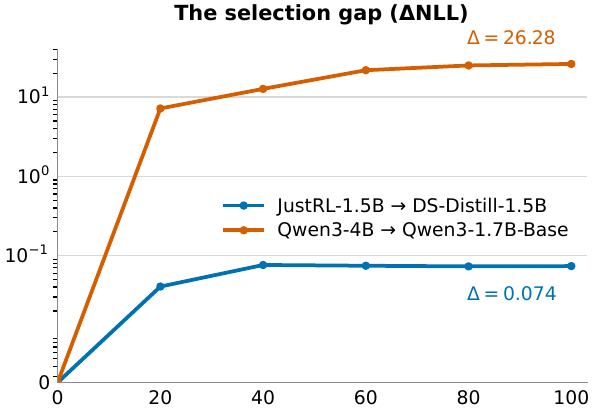}
    \captionsetup{font=small}
    \caption{
        \textbf{Selection gap before and after training.}
        Both students learn to discriminate teacher-preferred responses, with the Qwen3-4B setting showing overfitting.
    }
    \label{fig:selection_gap_comparison}
    \vspace{-5mm}
\end{wrapfigure}

\textbf{The trained student discriminates teacher-preferred responses.}
We measure how the trained student scores a fixed set of $S_{\mathrm{INIT}}$ responses.
We compare the 10\% of responses the teacher prefers most with the 10\% it prefers least, as ranked by teacher advantage.
For each response, $\Delta\text{NLL} = \text{NLL}_{S_t} - \text{NLL}_{S_0}$ measures the change in the student's NLL during training.
The \textit{selection gap} $\Gamma$ is the mean $\Delta\text{NLL}$ of the least-preferred 10\% minus that of the most-preferred 10\%.
A positive $\Gamma$ indicates that training favors the teacher's most-preferred responses over its least-preferred ones.

Figure~\ref{fig:selection_gap_comparison} shows whether this degradation reflects learned teacher preference rather than random drift. Both students increase their discrimination of teacher-preferred responses: JustRL's gap grows modestly from near zero to 0.07, while the Qwen3-4B setting's gap increases dramatically to approximately 26.
Combined with the endpoint generation table, this reveals that the student in the Qwen3-4B setting has \textit{overfit} to the subset of $S_{\mathrm{INIT}}$ responses that receive high teacher advantage but exhibit pathological behaviors.

\section{Diagnosing and Mitigating Reward Hacking}

To address reward hacking, prior work identifies which training samples exploit the reward model, then removes or reweights them during optimization~\citep{liu2024rrm,beigi2026adversarial,zhang2024policy}. Following this strategy, we first diagnose which $S_{\mathrm{INIT}}$ responses the teacher assigns high advantage to, then test whether masking pathological responses or reinitializing from a cleaner distribution can prevent the collapse.

\subsection{The Teacher's Reward Signal Can Be Misaligned with Response Quality}
\label{sec:teacher_selection}

\begin{figure}[t]
    \centering
    \includegraphics[width=0.95\textwidth]{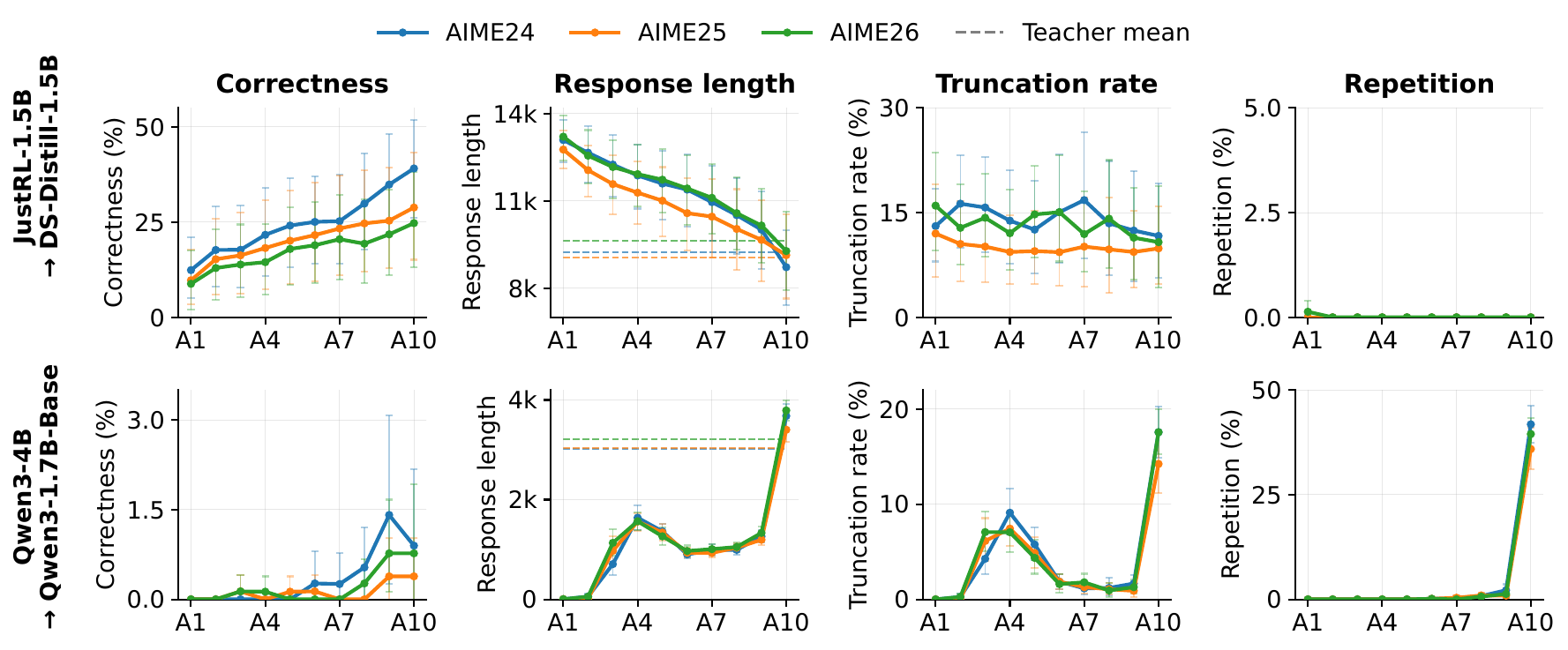}
    \caption{
        \textbf{Teacher advantage over $S_{\mathrm{INIT}}$ responses from A1 (lowest advantage) to A10 (highest advantage).}
        The JustRL teacher (top) assigns higher advantage to correct and shorter responses; the Qwen3-4B teacher (bottom) assigns higher advantage to overlong, repetitive ones.
        Dashed lines mark the teacher's own mean length.
    }
    \label{fig:teacher_preference_bias}
\end{figure}

To characterize the responses favored by the teacher before training, we rank all sampled $S_{\mathrm{INIT}}$ responses by their mean teacher advantage and divide them into ten equally sized groups, A1--A10.
Each group contains 10\% of the responses, with A1 containing the lowest-advantage responses and A10 the highest.
Figure~\ref{fig:teacher_preference_bias} reports correctness, response length,
generation-cap hit rate, and severe-repetition rate across advantage deciles.
In the JustRL setting, teacher advantage is aligned with response quality:
correctness increases toward A10, response length decreases toward the
teacher's own average, and no severe repetition is detected.
The Qwen3-4B setting shows the opposite pattern.
The highest-advantage responses are not the most reliable ones: their correctness remains close to zero, whereas their length, truncation rate, and repetition rate increase substantially across advantage deciles.
Although these pathological
responses are rare under $S_{\mathrm{INIT}}$, the teacher assigns them high
advantage, causing OPD to amplify them during training.
We also find similar failure pattern when the teacher is replaced as Qwen3-8B and Qwen3-30B-A3B.
Appendix~\ref{app:high_advantage_examples} presents three high-advantage
$S_{\mathrm{INIT}}$ responses that exhibit excessive length and mechanical repetition.

\begin{figure}[t]
    \centering
    \includegraphics[width=0.95\textwidth]{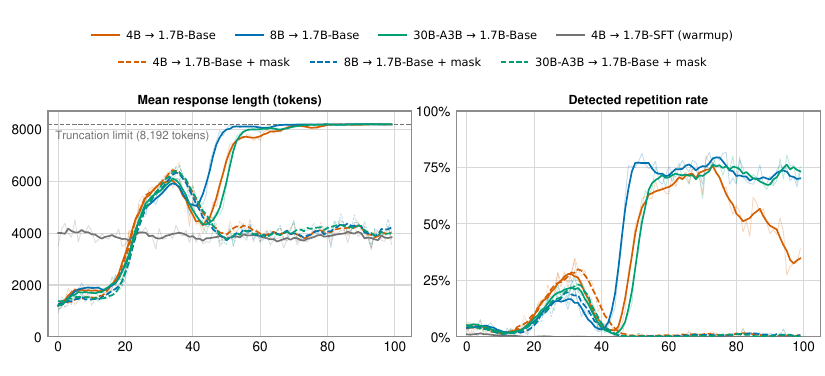}
    \caption{
        \textbf{Student-rollout characteristics during training.}
        All the masking settings recover from the collapse, whereas Qwen3-4B-warmup avoids it from the beginning.
    }
    \label{fig:candidate_pool_interventions}
\end{figure}

\subsection{Candidate-Pool Interventions Mitigate Reward Hacking}
\label{sec:5_2}

\begin{wraptable}{r}{0.49\textwidth}
    \centering

    \vspace{-2mm}
    \footnotesize
    \setlength{\tabcolsep}{2pt}
    \renewcommand{\arraystretch}{1.15}
    \caption{
        \textbf{Masking improves average accuracy across teachers.}
        Accuracy (\%) and the four-benchmark mean; teachers are Qwen3 models,
        and Base/SFT denote Qwen3-1.7B initializations.
    }
    \label{tab:candidate_pool_interventions}
    \resizebox{\linewidth}{!}{%
    \begin{tabular}{lrrrrr}
        \toprule
        \multirow{2}{*}{\textbf{Setting}} & \textbf{AMC}
        & \multicolumn{3}{c}{\textbf{AIME}} & \multirow{2}{*}{\textbf{Avg.}} \\
        \cmidrule(lr){2-2}\cmidrule(lr){3-5}
        & \textbf{23} & \textbf{24} & \textbf{25} & \textbf{26} & \\
        \midrule
        Base (pre-OPD) & 2.41 & 1.67 & 0.00 & 1.67 & 1.44 \\
        \midrule
        4B $\rightarrow$ Base & 29.82 & 3.33 & 3.33 & 4.17 & 10.16 \\
        \quad + mask & 34.64 & 8.33 & 5.83 & 4.17 & 13.24 \\
        \midrule
        8B $\rightarrow$ Base & 28.31 & 5.00 & 3.33 & 5.00 & 10.41 \\
        \quad + mask & 32.83 & 5.00 & 2.50 & 3.33 & 10.92 \\
        \midrule
        30B-A3B $\rightarrow$ Base & 28.61 & 6.67 & 2.50 & 4.17 & 10.49 \\
        \quad + mask & 33.13 & 10.00 & 4.17 & 5.00 & 13.07 \\
        \midrule
        4B $\rightarrow$ SFT$^{\dagger}$ (warmup) & 38.86 & 9.17 & 10.00 & 7.50 & 16.38 \\
        \bottomrule
    \end{tabular}%
    }
    \vspace{-2mm}
\end{wraptable}

The preference bias discussed previously is
already visible in responses sampled from $S_{\mathrm{INIT}}$, before any OPD
update takes place.
This suggests that the collapse may depend on which
responses are supplied to the teacher for supervision. To test this hypothesis,
we intervene on the candidate pool in the Qwen3-4B setting while keeping the
teacher, training data, and optimization recipe fixed.
We consider two interventions:
(1) The \textbf{Qwen3-4B-masked setting}: masking the loss on responses that reach the generation length limit.
(2) The \textbf{Qwen3-4B-warmup setting}: changing only the student initialization, from Qwen3-1.7B-Base to an SFT-initialized student released by \citet{li2026rethinkingopd}.
Appendix~\ref{app:intervention_details} provides the training details for both interventions.
\begingroup
\renewcommand{\thefootnote}{\fnsymbol{footnote}}
\footnotetext[2]{Warmup uses the Qwen3-1.7B-SFT checkpoint released by
\citet{li2026rethinkingopd}.}
\endgroup

\textbf{Masking recovers from collapse, while warmup avoids it.}
Figure~\ref{fig:candidate_pool_interventions} shows how response length, truncation, and repetition evolve over the course of training.
The original Qwen3-4B setting eventually enters a persistent collapsed state, where responses become increasingly long, truncated, and repetitive.
For the Qwen3-4B-masked setting, the student model exhibits few abnormal behaviors in the early training phase, so its learning trajectory closely aligns with that of the original Qwen3-4B.
As the undesirable behaviors accumulate, the effect of the masking intervention becomes clear: the Qwen3-4B-masked setting recovers, while the original setting collapses.
On the other hand, the warmup setting never develops this failure mode.
Warmup has been adopted as a remedy for OPD collapse~\citep{li2026rethinkingopd}.
The SFT-initialized student rarely produces the pathological responses favored by the teacher, leaving fewer such responses for OPD to reinforce.
Together, these results suggest that the collapse is not irreversible;
it depends on whether abnormal responses continue to receive reinforcement during training.

\textbf{Both interventions improve final accuracy.}
Table~\ref{tab:candidate_pool_interventions} extends the accuracy comparison to Qwen3-8B and Qwen3-30B-A3B teachers.
Masking improves the average accuracy over AMC23 and AIME24--26 by 3.08, 0.51, and 2.58 percentage points for the 4B, 8B, and 30B-A3B teachers, respectively.
These gains are not uniform across benchmarks: with the 8B teacher, AIME24 accuracy is unchanged while AIME25 and AIME26 accuracy decrease from 3.33\% to 2.50\% and from 5.00\% to 3.33\%, respectively.
The 4B warmup setting starts from a stronger SFT student and achieves the best final average accuracy (16.38\%), exceeding the corresponding Base-initialized OPD setting by 6.22 points.
In the 4B masking setting, accuracy improves even though only approximately 10\% of the sampled responses retain nonzero loss after abnormal responses are masked.

\section{Related Work}
\label{sec:related_work}

\textbf{Distillation paradigms and capability transfer.}
Knowledge distillation matches teacher distributions \citep{hinton2015distilling} or trains on teacher sequences \citep{kim2016sequence}; for reasoning models, the supervision is teacher rationales and trajectories \citep{ho2023large, hsieh2023distilling, mukherjee2023orcaprogressivelearningcomplex, mitra2023orca2teachingsmall, guo2025deepseek}, which extend coverage beyond the base model \citep{yue2025does}.
OPD instead supervises student-generated trajectories at student-visited states \citep{gu2024minillm, agarwal2023gkd}, with extensions to
self-distillation \citep{zhao2026selfdistilledreasoneronpolicyselfdistillation, ye2026onpolicycontextdistillationlanguage, he2026selfdistillationzeroselfrevisionturns},
representation-level supervision \citep{yang2026oprdonpolicyrepresentationdistillation},
multiple teachers \citep{ma2026mopdmultiteacheronpolicydistillation, yang2026learningteachergeneralizedonpolicy, team2026kimi},
and cross-tokenizer transfer \citep{he2026simpleopdsimpletokenizeragnosticonpolicy, wang2026crosstokenizeronpolicydistillationbyteprefix}.
These broaden teacher supervision but retain the reliance on student rollouts, over which our runs mainly redistribute probability rather than expanding the student's observed solvable set.

\textbf{Failure modes and stabilization of OPD.}
OPD failures arise from mismatches among student rollouts, teacher supervision, and token-level objectives.
\citet{li2026rethinkingopd} show that a stronger teacher need not help under reasoning-pattern mismatch; \citet{fu2026revisiting} and \citet{zhu2026facesonpolicydistillationpitfalls} identify unreliable guidance on student prefixes and bias from approximate reverse-KL updates.
Remedies include off-policy warm starts, data selection, stop-gradient objectives, trust regions \citep{xing2026trustregiononpolicydistillation}, and prefix correction \citep{jiang2026trajectoryrefineddistillation}.
Rollout-signal feedback creates another failure class: \citet{wang2026demystifyingonpolicydistillationroles} identify signal-correctness mismatch and length exploitation, while \citet{luo2026demystifyingopdlengthinflation} link length inflation to truncation and repetition collapse, motivating clipping, reference regularization, rollout mixing, and length control \citep{he2026simpleopdsimpletokenizeragnosticonpolicy, heo2026onpolicydeltadistillation}.
We instead view OPD as selecting among student-accessible behaviors by teacher preference, and analyze how biases in that preference affect training.

\textbf{Rollout pathologies in on-policy post-training.}
On-policy optimization trains only on trajectories sampled from the current model and is therefore exposed to the pathologies of its own rollout distribution.
For RLVR, pass@k and likelihood analyses show that training mainly redistributes probability over solution paths already accessible to the base model \citep{yue2025does}, and related studies find that reinforcement learning can amplify behaviors acquired during pretraining \citep{zhao2025echochamberrlposttraining} and reduce the diversity of reasoning trajectories \citep{dang2025assessing}.
Rollouts can further carry degenerate patterns such as malformed formats, excessive length, truncation, and self-reinforcing repetition. Repetition can be reinforced by token co-occurrence effects \citep{yan2024understanding}, while RLVR systems counter rollout pathologies with format rewards \citep{guo2025deepseek}, length-debiased advantage estimation \citep{liu2025understandingr1zeroliketrainingcritical}, and dynamic sampling with overlong filtering \citep{yu2025dapoopensourcellmreinforcement}.
OPD inherits this dependence on student rollouts but evaluates them through a teacher-derived signal; we show that degenerate student behaviors can bias the induced preference itself and thereby become preferentially retained during training.

\section{Conclusion}
\label{sec:conclusion}

We interpret on-policy distillation as an RL-like process in which the teacher acts as an implicit reward model for the student's own responses.
Across our experiments on competition mathematics, OPD improves sampling efficiency without expanding the observed solvable set.
Its gains primarily come from making correct responses already accessible to the initial student easier to sample.
This view also helps explain why OPD can collapse despite normal convergence.
In the collapsed setting, the student increasingly follows the teacher's preferences while producing longer and more repetitive responses.
These responses receive high teacher advantage even before training, although the teacher rarely generates them itself.
Masking the loss on truncated responses and initializing the student from an SFT checkpoint mitigate this failure while keeping the teacher fixed.

\section*{Limitations}

Computational constraints limit our experiments to small models and mathematical reasoning tasks, where prior work has reported clear cases of OPD success and collapse. Whether our findings extend to larger models or other domains remains to be studied.
Our interventions are designed specifically to address the overlong and repetitive generations observed in our experiments. We use them mainly to test our explanation of OPD collapse, rather than propose them as general solutions for OPD training.
Finally, our coverage results are based on a finite number of samples. Although we use additional sampling and manual inspection to check apparently new solutions, we cannot rule out that larger sampling budgets or different decoding settings would uncover additional correct responses from the initial model.

\section*{AI Use Statement}

We used generative AI tools to polish the manuscript, assist with writing
portions of the experimental code, and design the cartoon characters in the
main figure. The authors take responsibility for the final manuscript,
experimental implementation, and figures.

\section*{Ethics Statement}

This work studies the reasoning behavior and training dynamics of language
models using publicly available models and mathematical reasoning datasets.
We do not identify ethical concerns specific to the experiments presented here
beyond those associated with the underlying models and datasets.

\section*{Reproducibility Statement}

All models and datasets used in this work are publicly available, either
through open releases or accompanying prior publications. The appendix
describes the training configurations, evaluation protocols, and analysis
procedures. We plan to release all evaluation and analysis scripts to
facilitate reproduction of our results.

\bibliography{references}
\bibliographystyle{plainnat}

\newpage
\appendix

\section{Experimental Details}

\subsection{Training Settings}
\label{app:training_details}

We conduct three main experiments to study these phenomena.
The JustRL and DeepScaleR settings share the same student initialization and training data, while the Qwen3-4B
setting uses a base Qwen3 student and OpenThoughts3 prompts. The differences between the settings are summarized in Table~\ref{tab:appendix_training_protocol}.

\begin{center}
    \centering
    \small
    \setlength{\tabcolsep}{2.5pt}
    \captionof{table}{\textbf{Training settings used in the main experiments.}
    The teacher is fixed throughout OPD training.}
    \label{tab:appendix_training_protocol}
    \begin{tabular}{@{}lp{0.215\textwidth}p{0.235\textwidth}p{0.195\textwidth}cc@{}}
        \toprule
        Setting & Teacher & Student initialization & Training data & Length limit & LR \\
        \midrule
        JustRL & JustRL-1.5B
        & DeepSeek-R1-Distill-Qwen-1.5B
        & DeepMath-103K, level $\geq 6$ & 16,384 & $10^{-5}$ \\
        DeepScaleR & DeepScaleR-1.5B-Preview
        & DeepSeek-R1-Distill-Qwen-1.5B
        & DeepMath-103K, level $\geq 6$ & 16,384 & $10^{-5}$ \\
        Qwen3-4B & Qwen3-4B
        & Qwen3-1.7B-Base
        & OpenThoughts3 OPD 30K & 8,192 & $10^{-5}$ \\
        \bottomrule
    \end{tabular}
\end{center}

Each setting runs for 100 rollout iterations.
At each iteration, the student samples 8 responses for each of 32 prompts at
temperature $1.0$ and top-$p$ $1.0$, producing 256 responses and two optimizer
updates with global batch size 128. We use Adam with $\beta_1=0.9$,
$\beta_2=0.98$, weight decay $0.1$, no learning-rate warmup, PPO clipping
$\epsilon=0.2$, and an approximate reference-KL coefficient of $0.001$ toward
the fixed initialization $S_{\mathrm{INIT}}$.

The loss mask covers response tokens only.
Each sample is first averaged over its unmasked response tokens, and the resulting
values are then averaged over the global batch, so that long and short responses
contribute equally.
All runs use eight NVIDIA H100 GPUs.
Checkpoints are named by rollout count: the
initialization $S_{\mathrm{INIT}}$ together with iterations 20, 40, 60, 80, and
100, where iteration 100 gives $S_{\mathrm{OPD}}$.

\subsection{DeepScaleR Teacher-Temperature Pilot}
\label{app:deepscaler_temperature_pilot}

We ran two small temperature studies before fixing the DeepScaleR setting.
First, Table~\ref{tab:deepscaler_teacher_temperature} evaluates the frozen
DeepScaleR-1.5B-Preview teacher itself. For every temperature, we draw one
response per problem at top-$p$ $1.0$ with an 8,192-token limit and score it
with the verl verifier. The macro average weights AMC23, AIME24, and AIME25
equally despite their different problem counts.

\begin{center}
    \centering
    \small
    \setlength{\tabcolsep}{5.0pt}
    \captionof{table}{\textbf{Standalone teacher accuracy depends on the generation-sampling temperature.}
    Accuracy is in percent; Macro gives equal weight to the three benchmarks, and bold marks each column maximum.}
    \label{tab:deepscaler_teacher_temperature}
    \begin{tabular}{@{}lrrrr@{}}
        \toprule
        Temperature & AMC23 & AIME24 & AIME25 & Macro \\
        \midrule
        0.1 & 66.27 & 23.33 & \textbf{30.00} & 39.87 \\
        0.5 & \textbf{67.47} & 30.00 & 26.67 & 41.38 \\
        0.75 & 65.06 & \textbf{36.67} & 23.33 & \textbf{41.69} \\
        1 & 56.63 & 20.00 & 10.00 & 28.88 \\
        1.5 & 0.00 & 0.00 & 0.00 & 0.00 \\
        \bottomrule
    \end{tabular}
\end{center}

\begin{figure}[t]
    \centering
    \includegraphics[width=\textwidth]{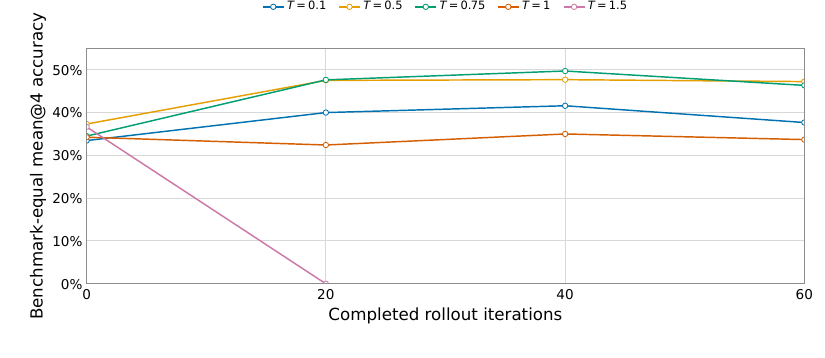}
    \caption{
        \textbf{Intermediate teacher-scoring temperatures give the strongest
        single-run student trajectories in the pilot.}
        Benchmark-equal student mean@4 accuracy is plotted against completed
        rollout iterations for the five teacher-scoring temperatures.
    }
    \label{fig:deepscaler_temperature_pilot}
\end{figure}

The training pilot in Figure~\ref{fig:deepscaler_temperature_pilot} instead
varies the temperature applied to the teacher logits when scoring student
rollouts. Each temperature corresponds to one training run. At each displayed
checkpoint, we evaluate the student with four responses per problem sampled at
temperature $1.0$, top-$p$ $1.0$, and a 16,384-token limit; we average within
each problem and then give the three benchmarks equal weight. The $T=1.5$ run
has observations only through iteration 20.

\FloatBarrier

These pilots support only a working choice, not an optimum. In the standalone
teacher study, $T=0.75$ has macro accuracy $41.69\%$, just $0.31$ percentage
points above $T=0.5$; weighting all 143 problems equally reverses the ordering
($50.35\%$ versus $51.05\%$). The teacher study uses one sample per problem,
and every training curve is a single run with unequal coverage at $T=1.5$.
Together with the strongest observed training point at $T=0.75$, these results
motivated adopting $0.75$ as the working temperature, but do not establish
that it is optimal.

\subsection{Evaluation Protocol}
\label{app:eval_protocol}

We use two evaluation protocols, and they differ only in how many responses are
drawn per problem. Both sample at top-$p$ $1.0$, use the same generation length
limit as training, and serialize prompts with the template described in
Appendix~\ref{app:prompts}. The sampling temperature follows the model being
evaluated rather than the protocol. In the JustRL setting and the DeepScaleR setting, $S_{\mathrm{INIT}}$
and $S_{\mathrm{OPD}}$ are both sampled at $T=1.0$, matching the rollout
temperature used in training.

\paragraph{Routine evaluation.}
Endpoint and per-checkpoint accuracies are reported as mean@4: we draw four
responses per problem, and calculate average accuracy among all the responses. This protocol is
used for the accuracy columns in Section~\ref{sec:5_2} and for the accuracy
traces over training.

\paragraph{Large-pool evaluation.}
The coverage and \passk{k} analyses in Section~\ref{sec:opd_phenomena} require a
much larger budget. For each model and each of the 90 AIME24--26 problems we
draw $n=256$ responses. This pool supports \passk{k} for
$k\in\{1,2,4,8,16,32,64,128,256\}$ and the per-problem success rate
$\hat p_m(q)$, defined as the fraction of verifier-correct responses among the
256 samples.

\paragraph{Conditional extension for the coverage audit.}
For the audit in Appendix~\ref{app:coverage_records} we extend sampling only
where it can change the conclusion. For every problem covered by
$S_{\mathrm{OPD}}$ but not by $S_{\mathrm{INIT}}$ at $n=256$, we draw a further
768 responses from $S_{\mathrm{INIT}}$, giving a conditional pool of 1{,}024
responses for that problem. The prompt, temperature, top-$p$, and length limit
are unchanged; only the number of samples increases.

\paragraph{The audit protocol is deliberately one-sided.}
Both halves of the audit act in a single direction, and we state this
explicitly rather than leave it to be inferred. The additional sampling is
spent only on problems that $S_{\mathrm{OPD}}$ covers and $S_{\mathrm{INIT}}$
does not; the reverse case---covered by $S_{\mathrm{INIT}}$ but not by
$S_{\mathrm{OPD}}$---never receives a larger $S_{\mathrm{OPD}}$ pool. Manual
review is applied to the same candidate list, where it reinstates verifier
misses among $S_{\mathrm{INIT}}$ responses and rejects invalid
$S_{\mathrm{OPD}}$ derivations. Relative to the unaudited 256-response pools,
the procedure can therefore only raise $S_{\mathrm{INIT}}$ coverage and only
lower $S_{\mathrm{OPD}}$ coverage, which is the direction that favors the
conclusion of Section~\ref{sec:opd_phenomena}.

We chose this design because a symmetric audit at equal budget would spend most
of it on problems whose status is not in question, and because the claim under
test is one-sided: a single surviving $S_{\mathrm{OPD}}$-only problem would
refute it, whereas additional $S_{\mathrm{INIT}}$-only problems would not
change it. The cost is that the audited partition in
Table~\ref{tab:appendix_coverage_sets} is not a symmetric measurement of the
two models. It should be read as an upper bound on $S_{\mathrm{INIT}}$
coverage and a lower bound on $S_{\mathrm{OPD}}$ coverage under this protocol,
so the empty $S_{\mathrm{OPD}}$-only column reports that no candidate survived
a search directed against it, not that the two supports were compared on equal
terms. The audited $S_{\mathrm{INIT}}$ and $S_{\mathrm{OPD}}$ totals are
therefore not on the same footing, and their difference overstates the true
gap.

A second consequence is that the candidate list differs between the two teacher
settings, so the two rows of Table~\ref{tab:appendix_coverage_sets} audit the
same $S_{\mathrm{INIT}}$ checkpoint under different budgets. We resolve this by
reporting a single $S_{\mathrm{INIT}}$ set: the union of the problems it is
found to solve under either audit. This is sound because $S_{\mathrm{INIT}}$ is
one checkpoint drawing on one shared pool of 256 responses per problem, so a
solution recovered in one audit is evidence about the checkpoint and not about
the teacher it was audited alongside. No counterpart exists on the
$S_{\mathrm{OPD}}$ side, where the two rows are different models. Pooling in
this way gives $S_{\mathrm{INIT}}$ the combined budget of both audits and so
widens the asymmetry above; it cannot create an $S_{\mathrm{OPD}}$-only
problem, since enlarging $S_{\mathrm{INIT}}$ can only remove entries from that
column.

\paragraph{Estimating \passk{k}.}
Reporting \passk{k} from a single group of $k$ responses is high variance. For a
problem with $c$ correct responses among $n$, we instead use the standard
unbiased estimator of the probability that a group of $k$ responses drawn
without replacement from the pool contains at least one correct response
\citep{chen2021evaluating},
\begin{equation}
    \passk{k}
    =
    \mathbb{E}_{q}
    \left[
        1-\frac{\binom{n-c_q}{k}}{\binom{n}{k}}
    \right],
    \label{eq:pass_at_k}
\end{equation}
evaluated at $n=256$ and averaged over problems with equal weight. We compute
the ratio as $\prod_{i=0}^{k-1}(n-c_q-i)/(n-i)$ to avoid large binomial
coefficients. At $k=n$ the estimator reduces to an indicator of whether the pool
contains any correct response, so \passk{256} equals the observed coverage of
that pool.

\paragraph{Correctness.}
Correctness is always recomputed from the response text; we do not reuse the
reward labels stored during rollout, which are produced by a different scorer
and under a different generation setting. All three benchmarks have
non-negative integer answers, and the verifier extracts a predicted integer with
the first rule that applies: the content of the last \verb|\boxed{...}| holding
an integer; otherwise the last integer introduced by an explicit answer cue such
as \texttt{final answer is}; otherwise the response as a whole, if it consists
of a single integer. A response with no extractable integer counts as incorrect.
Prediction and reference are compared after integer normalization, so the
verifier accepts no symbolic equivalence beyond leading-zero and whitespace
differences.

The second and third rules make the verifier permissive: a response can be
scored correct without ever emitting a boxed answer. This matters for the
coverage audit, where accepting an accidental hit from $S_{\mathrm{INIT}}$ would
weaken the claim it supports. We therefore inspect every
$S_{\mathrm{INIT}}$ recovery found in the extended pools by hand rather than
trusting the verifier alone, and report the outcome per problem in
Appendix~\ref{app:coverage_records}.

\subsection{Prompt Templates}
\label{app:prompts}

Training and evaluation examples contain one user message and no system
message. Each training example already includes the instruction
\texttt{\detokenize{Please reason step by step, and put your final answer within \boxed{}.}}
after the problem statement, separated by a newline in DeepMath and a space
in OpenThoughts3. We include the same instruction for mathematics evaluation
and serialize the message with the student's chat template. Routine
evaluation uses the same template as training. Below, \texttt{\{QUESTION\}}
denotes the problem statement without this instruction.

\paragraph{JustRL and DeepScaleR settings.}
Both use the DeepSeek-R1-Distill-Qwen-1.5B student template:
\begingroup
\expandafter\def\csname u8:｜\endcsname{\textbar}
\expandafter\def\csname u8:▁\endcsname{\rule{0.5em}{0.4pt}}
\begin{promptbox}{JustRL / DeepScaleR Prompt \textbar\ DeepSeek-R1-Distill-Qwen-1.5B}
\fontsize{7}{9}\selectfont
\begin{verbatim}
<｜begin▁of▁sentence｜><｜User｜>{QUESTION}
Please reason step by step, and put your final answer within \boxed{}.<｜Assistant｜><think>
\end{verbatim}
\end{promptbox}
\endgroup
The assistant prefix ends with \texttt{<think>} followed by a newline,
leaving the reasoning block open for generation.

\begin{samepage}
\paragraph{Qwen3-4B setting.}
We use the Qwen3-1.7B student template with thinking mode disabled.
Specifically, we prefill the assistant turn with an empty reasoning block,
\texttt{<think>} followed by two newlines, \texttt{</think>}, and two more
newlines. Generation therefore starts after the closed block. The serialized
prompt is:
\begin{promptbox}{Qwen3-4B Setting Prompt \textbar\ Qwen3-1.7B \textbar\ Thinking Disabled}
\fontsize{7}{9}\selectfont
\begin{verbatim}
<|im_start|>user
{QUESTION} Please reason step by step, and put your final answer within \boxed{}.<|im_end|>
<|im_start|>assistant
<think>

</think>

\end{verbatim}
\end{promptbox}
Disabling thinking mode changes this assistant prefix; the user instruction
to reason step by step remains in the prompt.
\end{samepage}

\paragraph{Teacher scoring.}
After the student generates a response, we feed the full token sequence---the
serialized prompt and the student-generated response---into the teacher in a
single forward pass to obtain the next-token distributions at the response
positions for distillation. The teacher generates no new tokens
(\texttt{max\_new\_tokens=0}). We use the student's original token sequence
directly, without reapplying a chat template or adding any instructions.

\FloatBarrier

\section{Observed Solvable Sets}
\label{app:coverage_records}

We first construct candidate solvable sets from problems with at least one
verifier-correct response in the original 256-response pools. We then refine
these sets through targeted sampling and manual review: invalid derivations
are excluded, and valid solutions missed by the automatic verifier are
reinstated. Table~\ref{tab:appendix_coverage_sets} reports the resulting
audited partition of all 90 AIME24--26 problems under each teacher.
The four columns form a disjoint partition: covered by both
$S_{\mathrm{INIT}}$ and $S_{\mathrm{OPD}}$, covered only by one of them, or
covered by neither. Problem numbers are listed explicitly within each dataset.

The audit combines additional sampling with review of the existing outputs.
We take every problem that is covered
by $S_{\mathrm{OPD}}$ but not by $S_{\mathrm{INIT}}$ among 256 responses and
extend the $S_{\mathrm{INIT}}$ pool to 1,024 responses. This recovers seven
cases: JustRL AIME25-025 and AIME26-010, and DeepScaleR AIME25-014,
AIME25-023, AIME25-025, AIME26-010, and AIME26-022. Manual review also checks the validity of
$S_{\mathrm{OPD}}$ derivations among the candidate problems, including cases
already recovered by additional sampling. The JustRL AIME25-020 and
DeepScaleR AIME25-014 and AIME25-025 derivations are invalid despite reaching
the correct final answer. The latter two problems are therefore assigned to
the $S_{\mathrm{INIT}}$-only set.

For JustRL AIME26-022 and AIME26-027, the automatic verifier records no
$S_{\mathrm{INIT}}$ hit in the 1,024-response pools. Manual review identifies
valid $S_{\mathrm{INIT}}$ solutions in the existing outputs and corrects these
verifier misses; it does not use a separate batch of generations. These cases
are assigned to the shared set.
Table~\ref{tab:appendix_coverage_audit} distinguishes recoveries by additional
sampling from corrections to automatic verification.

\begin{center}
\captionof{table}{\textbf{Audited observed-solvable-set partitions.}
Each cell gives the number of problems followed by their three-digit suffixes.}
\label{tab:appendix_coverage_sets}
\centering
\scriptsize
\setlength{\tabcolsep}{2.2pt}
\begin{tabular}{llp{0.31\textwidth}p{0.12\textwidth}p{0.10\textwidth}p{0.20\textwidth}}
\toprule
Teacher & Data & Both & $S_{\mathrm{INIT}}$ only & $S_{\mathrm{OPD}}$ only & Neither \\
\midrule
JustRL & AIME24
& \textbf{24}: 001--007, 009, 010, 013--022, 025--029
& \textbf{1}: 023 & \textbf{0}: --
& \textbf{5}: 008, 011, 012, 024, 030 \\
JustRL & AIME25
& \textbf{18}: 001--006, 008, 009, 012, 016, 017, 019, 021, 022, 024, 025, 027, 029
& \textbf{6}: 010, 013, 014, 023, 028, 030 & \textbf{0}: --
& \textbf{6}: 007, 011, 015, 018, 020, 026 \\
JustRL & AIME26
& \textbf{23}: 001--008, 010, 012--014, 016, 018--027
& \textbf{1}: 011 & \textbf{0}: --
& \textbf{6}: 009, 015, 017, 028--030 \\
\midrule
DeepScaleR & AIME24
& \textbf{23}: 001--007, 009, 010, 013--022, 025--028
& \textbf{2}: 023, 029 & \textbf{0}: --
& \textbf{5}: 008, 011, 012, 024, 030 \\
DeepScaleR & AIME25
& \textbf{18}: 001--006, 008, 009, 012, 016, 017, 019, 021--024, 027, 029
& \textbf{6}: 010, 013, 014, 025, 028, 030 & \textbf{0}: --
& \textbf{6}: 007, 011, 015, 018, 020, 026 \\
DeepScaleR & AIME26
& \textbf{22}: 001--008, 010, 012--014, 016, 018--026
& \textbf{2}: 011, 027 & \textbf{0}: --
& \textbf{6}: 009, 015, 017, 028--030 \\
\bottomrule
\end{tabular}
\end{center}

\begin{center}
\captionof{table}{\textbf{Resolution of the raw $S_{\mathrm{OPD}}$-only cases.}
The third column records how a correct $S_{\mathrm{INIT}}$ response was
obtained. ``Extended sampling'' means the verifier accepted a response drawn
while extending the pool to 1,024. ``Manual review'' means a correct response
was already present and had been missed by the verifier, so no additional
generation was needed. ``None'' means neither route produced one.}
\label{tab:appendix_coverage_audit}
\centering
\small
\setlength{\tabcolsep}{3.5pt}
\begin{tabular}{@{}llp{0.22\textwidth}p{0.17\textwidth}p{0.27\textwidth}@{}}
\toprule
Teacher & Problem & $S_{\mathrm{INIT}}$ solution found by & $S_{\mathrm{OPD}}$ derivation & Resolution \\
\midrule
JustRL & AIME25-020 & None & Invalid & Removed as an accidental verifier hit \\
JustRL & AIME25-025 & Extended sampling & Valid & Shared \\
JustRL & AIME26-010 & Extended sampling & Valid & Shared \\
JustRL & AIME26-022 & Manual review & Valid & Shared \\
JustRL & AIME26-027 & Manual review & Valid & Shared \\
\midrule
DeepScaleR & AIME25-014 & Extended sampling & Invalid & $S_{\mathrm{INIT}}$ only \\
DeepScaleR & AIME25-023 & Extended sampling & Valid & Shared \\
DeepScaleR & AIME25-025 & Extended sampling & Invalid & $S_{\mathrm{INIT}}$ only \\
DeepScaleR & AIME26-010 & Extended sampling & Not needed & Shared \\
DeepScaleR & AIME26-022 & Extended sampling & Valid (two reviews) & Shared \\
\bottomrule
\end{tabular}
\end{center}

These sets describe the responses observed under the stated sampling and
manual-audit protocol. They do not establish equality of model support or
zero probability for any response under $S_{\mathrm{INIT}}$.

\FloatBarrier

\section{Min-10NN Distance}
\label{app:response_geometry}

Min-10NN measures how tightly responses to the same problem concentrate in
surface form. For two decoded responses $a$ and $b$, we use character-level
normalized Levenshtein distance
\begin{equation}
    d(a,b)=
    \begin{cases}
        0, & |a|=|b|=0,\\
        \operatorname{Lev}(a,b)/\max\{|a|,|b|\}, & \text{otherwise}.
    \end{cases}
\end{equation}

Let $R=\{r_i\}_{i=1}^{16}$ be 16 responses to one problem. For each response,
we compute its distance to the nearest other response,
\begin{equation}
    \delta_i=\min_{j\ne i}d(r_i,r_j).
\end{equation}
After sorting these values as
$\delta_{(1)}\leq\cdots\leq\delta_{(16)}$, Min-10NN is
\begin{equation}
    \operatorname{Min\text{-}10NN}(R)
    =\frac{1}{10}\sum_{i=1}^{10}\delta_{(i)}.
\end{equation}
A smaller value means that the ten most tightly clustered responses are more
similar. When 32 responses are available, we sample 16 without replacement,
average Min-10NN over 100 repetitions, and then average across problems. The
metric captures character-level concentration only; it does not measure
semantic equivalence or response quality.

\FloatBarrier

\section{Response Case Study}
\label{app:response_excerpts}

Figure~\ref{fig:qualitative_response_examples} gives the examples of the
excerpts condensed in Figure~\ref{fig:main_teaser}b. All four responses come
from the Qwen3-4B setting before any OPD update: (a) is a free generation of
the teacher, and (b)--(d) are sampled from $S_{\mathrm{INIT}}$, with (c) and
(d) drawn from the responses the teacher ranks as most preferred under the
decile partition of Section~\ref{sec:teacher_selection}.

\begin{figure*}[t]
    \centering
    \begingroup
    \definecolor{responsehealthyframe}{HTML}{3C7D73}
    \definecolor{responsehealthyfill}{HTML}{F1F8F6}
    \definecolor{responsecollapseframe}{HTML}{B85245}
    \definecolor{responsecollapsefill}{HTML}{FCF3F1}
    \setlength{\fboxsep}{5pt}
    \setlength{\fboxrule}{0.55pt}
    \newcommand{\responseexample}[5]{%
        \begin{minipage}[t]{0.485\textwidth}
            \fcolorbox{#1}{#2}{%
                \parbox[t][25mm][t]{\dimexpr\linewidth-2\fboxsep-2\fboxrule\relax}{%
                    \raggedright
                    \scriptsize
                    \textbf{#3}\par
                    {\fontsize{6.8}{7.5}\selectfont\textbf{#4}}\par
                    \vspace{1.5pt}
                    #5
                }%
            }
        \end{minipage}%
    }

    \responseexample{responsehealthyframe}{responsehealthyfill}
        {(a) Teacher: coherent solution}{correct; 842 tokens}{%
        ``Number of \textbf{grand prize} outcomes: 1. \ldots\ Number of
        \textbf{prize} outcomes ($\geq 2$ matches): 115. \ldots
        $P(\text{Grand Prize}\mid\text{Prize})=1/115$.
        Thus, $m+n=1+115=116$.''}
    \hfill
    \responseexample{responsehealthyframe}{responsehealthyfill}
        {(b) $S_{\mathrm{INIT}}$: coherent solution}{correct; 771 tokens}{%
        ``Jen has exactly 2 of the chosen 4 numbers:
        $C(4,2)C(6,2)=90$. \ldots\ Exactly 3: 24; all 4: 1.
        So, the total number of ways to win a prize is
        $90+24+1=115$. \ldots
        $P(\text{grand prize}\mid\text{prize})=1/115$.
        \ldots\ \fbox{116}.''}

    \vspace{4pt}

    \responseexample{responsecollapseframe}{responsecollapsefill}
        {(c) $S_{\mathrm{INIT}}$, teacher-preferred: repeated reconsideration}{incorrect; 7,311 tokens}{%
        ``$17{,}499{,}840/661{,}680=26.5$.
        \textbf{Wait, this is incorrect.} The calculation is not correct.
        Instead, we should compute the number of valid sequences. Let's go
        back. \ldots\ But this is not correct. \textbf{Wait, for the last word}
        to be GY, \ldots\ \textbf{Wait, this is not correct.}''}
    \hfill
    \responseexample{responsecollapseframe}{responsecollapsefill}
        {(d) $S_{\mathrm{INIT}}$, teacher-preferred: mechanical repetition}{correct; 8,192 tokens; truncated}{%
        ``\textbf{Final Answer: \fbox{116}} \ldots\
        \textbf{Answer: \fbox{116}} \ldots\
        \textbf{Final Answer: \fbox{116}} \ldots\
        \textbf{Answer: \fbox{116}} \ldots''\par
        \vspace{2pt}
        The same answer line repeats 365 times until the generation length
        limit.}
    \endgroup
    \caption{
        \textbf{Illustrative response excerpts from the Qwen3-4B setting.}
        The teacher and a typical $S_{\mathrm{INIT}}$ response give coherent
        solutions, whereas the $S_{\mathrm{INIT}}$ responses the teacher most
        prefers repeatedly restart their reasoning or mechanically repeat an
        answer until the generation length limit. All four excerpts predate any
        OPD update. Ellipses mark omitted text. These examples illustrate
        behaviors and do not estimate their prevalence.
    }
    \label{fig:qualitative_response_examples}
\end{figure*}

\FloatBarrier

\section{High-Advantage Responses from the Initial Student}
\label{app:high_advantage_examples}

We present three responses sampled from Qwen3-1.7B-Base before any OPD
update and scored by the Qwen3-4B teacher. All three belong to the
highest-advantage group, A10, in the analysis of
Section~\ref{sec:teacher_selection} and illustrate excessive length and
mechanical repetition already present under $S_{\mathrm{INIT}}$.
All three are incorrect and contain no parseable final answer.

\paragraph{Reading the complete responses.}
Each response consists almost entirely of a repeated string. We specify its
complete content using exact repetition counts: $+$ denotes concatenation
without an added separator, and $\operatorname{repeat}(s,n)$ denotes $n$
consecutive copies of $s$. Quotation marks delimit strings and are not part
of the response; \texttt{\textbackslash n} denotes a newline. Unicode
characters are written as \texttt{\textbackslash uXXXX} escapes. No text is
omitted from these representations. Prompt mathematics below is
typeset for readability. Every prompt ends with: ``Please reason step by
step, and put your final answer within \texttt{\textbackslash boxed\{\}}.''

\subsection{Length-Cap Repetition}

\paragraph{Prompt (AIME24, \texttt{aime24\_001}).}
Every morning Aya goes for a $9$-kilometer-long walk and stops at a coffee
shop afterwards. When she walks at a constant speed of $s$ kilometers per
hour, the walk takes her 4 hours, including $t$ minutes spent in the coffee
shop. When she walks $s+2$ kilometers per hour, the walk takes her 2 hours
and 24 minutes, including $t$ minutes spent in the coffee shop. Suppose Aya
walks at $s+\frac{1}{2}$ kilometers per hour. Find the number of minutes the
walk takes her, including the $t$ minutes spent in the coffee shop.

\paragraph{Response (sample 65).}
\begin{quote}
\small\ttfamily
"\textbackslash uFFFD\textbackslash uFFFD\textbackslash u6001\ T 10T 50 10 1 10 10"\\
\normalfont $+\;\operatorname{repeat}($\texttt{"T 10"}$,\;2043)$
\end{quote}
The response mechanically repeats \texttt{T 10} until it reaches the
8,192-token generation cap. The two initial
\texttt{\textbackslash uFFFD} characters are present in the stored response.

\subsection{Severe Repetition without Reaching the Length Cap}

\paragraph{Prompt (AIME25, \texttt{aime25\_028}).}
Let the sequence of rationals $x_1,x_2,\ldots$ be defined such that
$x_1=\frac{25}{11}$ and
$x_{k+1}=\frac{1}{3}\left(x_k+\frac{1}{x_k}-1\right)$.
$x_{2025}$ can be expressed as $\frac{m}{n}$ for relatively prime positive
integers $m$ and $n$. Find the remainder when $m+n$ is divided by 1000.

\paragraph{Response (sample 110).}
\begin{quote}
\small
$\operatorname{repeat}($\texttt{".slfwith\textbackslash n"}$,\;2154)
+$ \texttt{".slfwith"}
\end{quote}
The response consists of repeated \texttt{.slfwith} lines and ends after
6,465 recorded tokens, showing that severe repetition can occur without
reaching the generation cap.

\subsection{Severe Repetition at the Length Cap}

\paragraph{Prompt (AIME25, \texttt{aime25\_013}).}
Alex divides a disk into four quadrants with two perpendicular diameters
intersecting at the center of the disk. He draws 25 more line segments
through the disk, drawing each segment by selecting two points at random
on the perimeter of the disk in different quadrants and connecting those
two points. Find the expected number of regions into which these 27 line
segments divide the disk.

\paragraph{Response (sample 58).}
\begin{quote}
\small\ttfamily
"\textbackslash u6CD5\ newVal89\textbackslash u6CD5\ newVal8\textbackslash u6CD5\ newVal8"\\
\normalfont $+\;\operatorname{repeat}($\texttt{"\textbackslash u6CD5\ newVal"}$,\;4091)$
\end{quote}
The response repeats the same fragment until it reaches the 8,192-token
generation cap, exhibiting both excessive length and severe repetition.

Together, these cases illustrate how excessively long and repetitive responses can receive
high relative teacher advantage already under $S_{\mathrm{INIT}}$.

\FloatBarrier

\section{Training Details for the Candidate-Pool Interventions}
\label{app:intervention_details}

Section~\ref{sec:5_2} compares three Qwen3-4B teacher settings. The baseline
and masked settings initialize the student from Qwen3-1.7B-Base. The warmup
setting instead initializes it from the Qwen3-1.7B-SFT checkpoint released by
\citet{li2026rethinkingopd} and uses that model as the fixed reference.
All three use OpenThoughts3 prompts, temperature $1.0$, an
8,192-token generation length limit, and the optimization settings in
Appendix~\ref{app:training_details}.

\paragraph{Loss masking for truncated or failed responses.}
The baseline trains on all sampled responses, including those that reach the
generation length limit. In the masked setting, we set the loss mask to zero
for truncated or failed responses; completed responses retain nonzero loss
masks and contribute to training. Truncated or failed responses remain in reward and
advantage normalization but contribute no policy, reference-KL, or entropy
gradient. The loss is first averaged over unmasked response tokens within
each sample and then over samples in the global batch.

\paragraph{Initialization from a released SFT checkpoint.}
The warmup setting directly loads the Qwen3-1.7B-SFT checkpoint released by
\citet{li2026rethinkingopd}; we do not perform an additional SFT stage ourselves.
It changes only the student initialization and fixed reference
from Qwen3-1.7B-Base to Qwen3-1.7B-SFT; the teacher, data, optimizer, sampling
protocol, and evaluation schedule remain fixed. Its higher initial accuracy
is part of the intervention: it tests whether supplying a better initial
candidate pool changes what OPD can amplify.

\FloatBarrier

\end{document}